\documentclass[runningheads]{llncs}
\usepackage{graphicx}
\usepackage{placeins}
\usepackage{booktabs}
\usepackage{multirow}
\usepackage{makecell}
\usepackage{cite}
\usepackage{epstopdf}
\usepackage{amsmath}

\begin{document}
\title{Improving Neural Network Classifier using Gradient-based Floating Centroid Method}
%
%
\author{Mazharul Islam\inst{1,\#} \and
Shuangrong Liu\inst{1,\#} \and
Lin Wang\inst{1,*} \and
Xiaojing Zhang\inst{1}}
\titlerunning{Gradient-based Floating Centroid Method}
\authorrunning{M. Islam et al.}
%
\institute{Shandong Provincial Key Laboratory of Network Based Intelligent Computing, University of Jinan, Jinan, 250022, China  \\
\email{*Corresponding Author: wangplanet@gmail.com}}
\maketitle              
\footnote{\# Both authors contribute equally to this article.}

\begin{abstract}
Floating centroid method (FCM) offers an efficient way to solve a fixed-centroid problem for the neural network classifiers. However, evolutionary computation as its optimization method restrains the FCM to achieve satisfactory performance for different neural network structures, because of the high computational complexity and inefficiency. Traditional gradient-based methods have been extensively adopted to optimize the neural network classifiers. In this study, a gradient-based floating centroid (GDFC) method is introduced to address the fixed centroid problem for the neural network classifiers optimized by gradient-based methods. Furthermore, a new loss function for optimizing GDFC is introduced. The experimental results display that GDFC obtains promising classification performance than the comparison methods on the benchmark datasets.
\end{abstract}

\section{Introduction}
In the machine learning field, supervised classification is a classical topic among scientists. There are a considerable number of classification methods has been proposed during past years, such as naive bayes \cite{Wang2010}, k-nearest neighbor\cite{Cunningham2010}, decision tree\cite{Quinlan86inductionof}, support vector machine \cite{Chua:2003:ECL:2840212.2840505}, and neural network \cite{7990591}. Among these methods, the neural network attracts substantial attention for addressing real-world classification problems \cite{Jiang2018,NIPS2018_8190,kamilaris_2018,Wibowo_2019} because of its capability of learning non-linear relationship from real-world data.

Conventionally, the classification process of the neural network is explained from the probabilistic perspective. The values of output neurons are considered as the probabilities that a sample belongs to different classes. Meanwhile, this process is also can be described from a geometric perspective. The neural network is viewed as a mapping function \textbf{\textit{f}}. Each class is coded as the unique binary string. As the input, the sample is mapped to a space by \textbf{\textit{f}}, in which classes are represent by different fixed points, referred to centroids. The binary string of each class describes the position of centroids. The mapped sample is attached to the closest centroid (class) in according to the distance. Therefore, methods acting on the output layer of the neural network, such as one-per-class, softmax\cite{10.1007/978-3-642-76153-9_28} and error-correcting output code (ECOC)\cite{Dietterich:1995:SML:1622826.1622834}, can also be viewed as the methods to distribute the centroids.

For the one-per-class, softmax and ECOC, they are widely used in different neural network models optimized by gradient-based optimization method, and obtains considerable successful stories\cite{Wibowo_2019, Wong2018}. However, for these fixed centroid methods, the fixed centroid problem (FCP)\cite{10.1007/978-3-540-73053-8_21, 7502076, Wang2012}, which refers to that the locations, labels, and number of centroids are prior set before the training, restrains their performance. Because the FCP results in the reduction of the size of the set consisting of optimal neural networks, and enlarges the complexity of optimization. Although the floating centroid method (FCM)\cite{Wang2012} affords a way to solve the fixed centroid problem, the evolutionary computation is adopted as its optimization method that prevents FCM to employ with the neural networks optimized by gradient-based optimization method.

Considering the facts mentioned above, gradient-based floating centroid method (GDFC) is proposed in this study. The GDFC absorbs the advantages of fixed centroids methods and floating centroids method, and affords a way to assist the neural network classifiers optimized by gradient-based optimization method to address the fixed centroid problem. For this study, the major contributions are introduced as follows:
\begin{itemize}
  \item  The gradient-based floating centroid method is proposed to tackle fixed-centroid problem for the neural network classifiers optimized by gradient-based methods.
  \item A new loss function, named centroid loss function, is proposed to maximize compactness of within-class and separability of between-class during training process.

\end{itemize}

 The proposed GDFC method is described in Section 2. Experiment is reported and results on benchmark datasets are analyzed in Section 3. We give the conclusion, and draw the future work in Section 4,.

\section{Methodology}
In this section, the framework of gradient-based floating centroid method is provide firstly. Subsequently, the centroid loss function is introduced. At the end of this section, we describe the optimization process of GDFC.

\subsection{Gradient-based Floating Centroid Framework}

\begin{figure}[!hbt]
\centering
\includegraphics[width = 13cm]{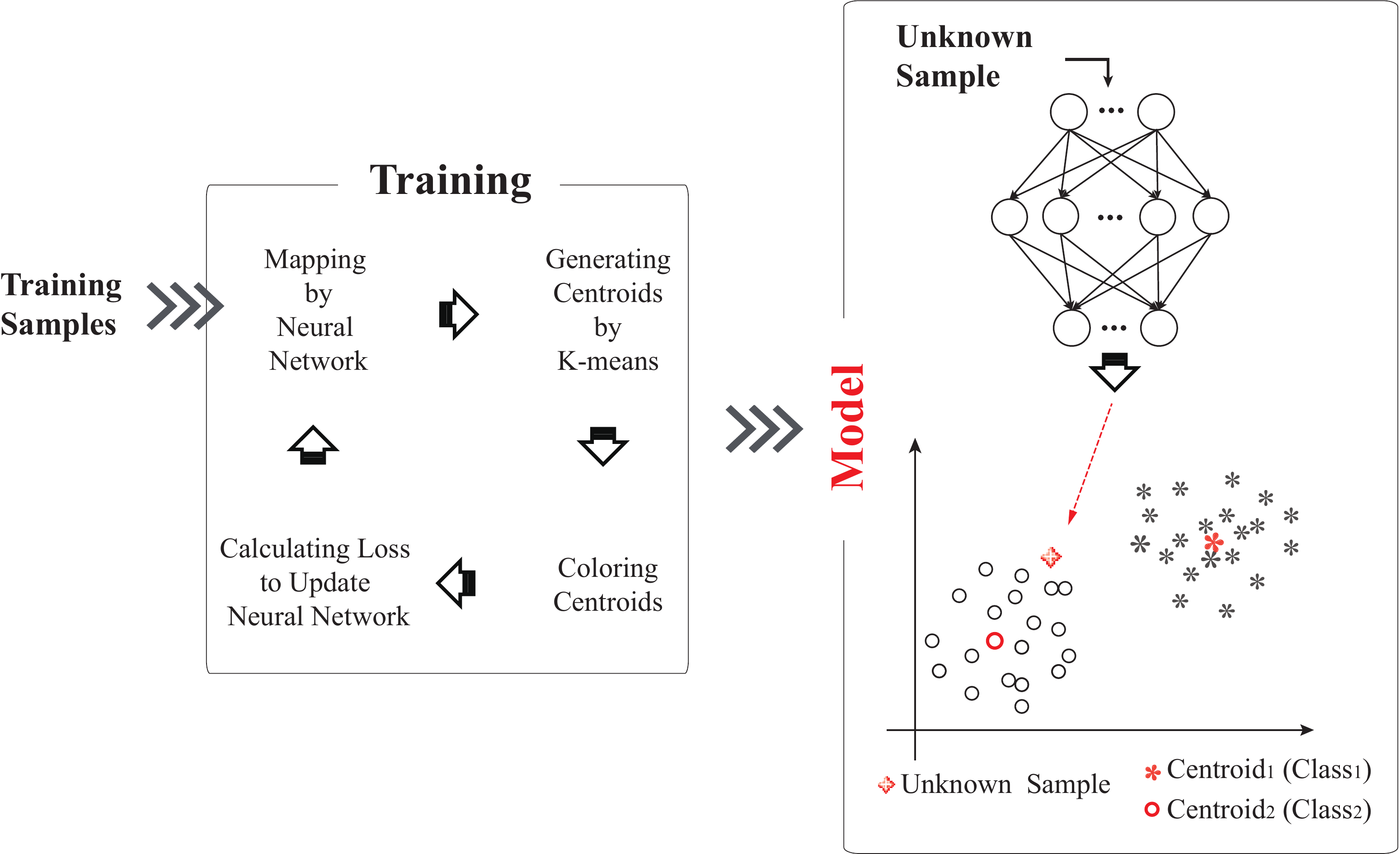}
\caption{Architecture of the GDFC}
\label{GDFCframework}
\end{figure}

The framework of the gradient-based floating centroid is shown in Fig. 1. The training part mainly includes 4 modules: mapping by neural network, generating centroids by K-means, coloring centroids, and calculating loss to update neural network. At first, the neural network maps the samples to the partition space. Afterwards, the centroids are generated in the partition space by using k-means algorithm. The classes' number can smaller than the number of centroids. Subsequently, these centroids are labeled by different classes. The labeling strategy refer to coloration process. In the coloration process, if the mapped samples which is represented by one class are the majority then the corresponding centroid is colored by that class. Besides, one class can be used to label more than one centroids. After that, the neural network is iteratively updated by the proposed centroid loss function which uses the distribution information of the centroids. In the optimization process, the centroid loss function has the ability to maximize compactness of within-class and separability of between-class simultaneously; thus clear decision boundaries exist among different clusters. Finally, an optimal neural network and centroids decided by this optimal neural network are obtained.

In the testing process, an unknown sample as input of the optimal neural network are mapped to the partition space. This unknown sample is assigned to the centroid with closest distance. For example, in Fig.1, the unknown sample is close to the centroid, which represents class 2. So, this sample categorized to class 2.

\subsection{Centroid Loss Function}
For the mapped samples in the partition space, the k-means algorithm is used to generate the centroids $C^{(k)} (k=1,2,...,K)$ by clustering these mapped samples. Then, for each mapped sample, two centroids are selected from K centroids:

The first centroid is one with minimum value of the $||\cdot ||_{2}$, and this centroid having same class as the mapped sample.

\begin{equation}
D_{\min }^{Self} =\arg \min _{C} ||\beta ^{(j)} -C^{(k)} (g)||_{2}
\end{equation}
where \emph{j=1,2,...,m}, $m$ is the number of samples, \emph{k=1,2,..,K}, $K$ is the number of centroids, $||\cdot ||_{2} $ represents the distance.
Note that we can obtain the centroid which has the same class and nearest distance to the mapped sample, denotes $C^{S} $. The second centroid is one with minimum value of the $||\cdot ||_{2}$, and the class of this centroid is different with the mapped sample.
\begin{equation}
D_{\min }^{Noself} =\arg \min _{C} ||\beta ^{(j)} -C^{(k)} (g)||_{2}
\end{equation}
Note that we can select the centroid of the different class nearest to the  mapped sample, denotes $C^{N} $. Since the target of GDFC is to put the points belongs to the same class closer and enlarge the distance among the points with different classes, thus, minimizing the $D_{\min }^{Self} $ as well as maximizing the $D_{\min }^{Noself} $ are expected. Adopting the method of stochastic gradient descent, which attempts to minimize the global error by updating the parameters of the neural network in an iterative process. Therefore, the loss function is listed as follow.
\begin{equation}
E=\frac{1}{2} \sum _{q=1}^{Q}\left[(\beta _{qj} -C_{qj}^{S} )^{2} -\xi \cdot (\beta _{qj} -C_{qj}^{N} )^{2} \right]
\end{equation}
$\beta _{qj} $ denotes the mapped value of the qth (q=1,2,..,Q) neuron in the output layer, Q represents the output neurons' number, and is also the dimension of the partition space. $\xi$ is a constant, which is used to adjust the weight between $D_{\min }^{Self} $ and $D_{\min }^{Nosolf} $. Besides, the gradient descent method is prone to over-fitting, so $L_{2}$ regularization is applied to decrease over-fitting. From the above, the centroid loss function is essential to make the following reformulation as,
\begin{equation}
E=\frac{1}{2} \sum _{q=1}^{Q}\left[ ((\beta _{qj} -C_{qj}^{S} )^{2} -\xi \cdot (\beta _{qj} -C_{qj}^{N} )^{2} )\right]+\frac{\lambda }{2}  \cdot \sum w^{2}
\end{equation}
Where $\lambda $ is the regularization parameter.

\subsection{Optimization}
Based on the gradient descent method, while following the back-propagation (BP) idea of training-error to iteratively update the parameters to obtain an optimal neural network. Without loss of generality, assuming that a feedforward neural network has L layers. $\eta$ is a global learning rate. From the L layer to the L-1 layer, the partial derivatives of the weights and biases are obtained respectively.
\begin{equation}
\begin{split}
\Delta w_{qh}^{(L-1)}
        &=-\eta \frac{\partial E}{\partial w_{qh}^{(L-1)} } \\
        &=-\eta \cdot \frac{\partial E}{\partial \beta _{qj} } \frac{\partial \beta _{qj} }{\partial z_{qj} } \frac{\partial z_{qj} }{\partial w_{qh}^{(L-1)} } -\lambda \cdot w_{qh}^{(L-1)}  \\
        &=\eta \left[\delta _{q}^{(L)} \alpha _{hj}^{(L-1)} -\lambda w_{qh}^{(L-1)} \right]
\end{split}
\end{equation}

\begin{equation}
{\Delta \theta _{q}^{(L)} =\eta \frac{\partial E}{\partial \theta _{q}^{(L-1)} } }
{{\mathop{}\limits^{}} {\mathop{}\limits^{}} {\mathop{}\limits^{}} =\eta \frac{\partial E}{\partial \beta _{qj} } \frac{\partial \beta _{qj} }{\partial \theta _{q}^{(L-1)} } } =-\eta \delta _{q}^{(L)}
\end{equation}

Note that, \begin{equation}
\delta _{q}^{(L)} =-\frac{\partial E}{\partial \beta _{qj} } \frac{\partial \beta _{qj} }{\partial z_{qj} } =(C_{qj}^{S} -\beta _{qj} )\cdot \sigma ^{'} (z_{qj} )-\xi  (C_{qj}^{N} -\beta _{qj} )\sigma ^{'} (z_{qj} )
\end{equation}
Where Q is neurons' number in L layer, H represents the neurons' number in the L-1 layer, $\Delta w_{qh}^{(L-1)} $ is the weight change value from the \emph{h}th (with h=1,2,...,H) neuron of the L-1 layer to the \emph{q}th (q=1,2,..,Q) neuron of the L layer, $\Delta \theta _{q}^{(L)} $ is the bias change value of the \emph{q}th (with q=1,2,..,Q) neuron in the L layer. $\sigma (\cdot )$ represents the activation function. $\beta _{qj} = \sigma (z_{qj} )=\sigma (\sum _{h=1}^{H}(w_{qh}^{(L-1)} \cdot \alpha _{hj}^{(L-1)} +\theta _{q}^{(L)})$, $\beta $ is the activation value of the L layer neurons, $\alpha $ is the activation value of the L-1 layer neurons.
Thus, the weights and biases between the L layer and the L-1 layer are updated as
\begin{equation}
w_{qh}^{(L-1)} (g+1)=w_{qh}^{(L-1)} {\rm (g)}+\Delta w_{qh}^{(L-1)}
\end{equation}
\begin{equation}
\theta _{q}^{(L-1)} (g+1)=\theta _{q}^{(L-1)} (g)+\Delta \theta _{q}^{(L-1)}
\end{equation}
From the L-1 layer to the L-2 layer, the partial derivatives of the weights and biases are obtained, respectively

\begin{equation}
\begin{split}
\begin{array}{l} {\Delta w_{hp}^{(L-2)} =-\eta \frac{\partial E}{\partial w_{hp}^{(L-2)} } } \\ {{\mathop{}\nolimits^{}} {\mathop{}\nolimits^{}} {\mathop{}\limits^{}} =-\eta \cdot \left[\frac{\partial E}{\partial \beta _{qj} } \frac{\partial \beta _{qj} }{\partial z_{qj} } \frac{\partial z_{qj} }{\partial \alpha _{hj}^{(L-1)} } \frac{\partial \alpha _{hj}^{(L-1)} }{\partial y_{hj}^{(L-1)} } \frac{\partial y_{hj}^{(L-1)} }{\partial w_{hp}^{(L-2)} } \right]-\eta \cdot \lambda \cdot w_{hp}^{(L-2)} } \\ {{\mathop{}\nolimits_{}} {\mathop{}\nolimits_{}} {\mathop{}\limits^{}} =\eta \cdot \left[\delta _{h}^{(L-1)} \alpha _{pj}^{(L-2)} -\lambda w_{hp}^{(L-2)} \right]} \end{array}
\end{split}
\end{equation}

\begin{equation}
{\Delta \theta _{h}^{(L-1)} =-\eta \frac{\partial E}{\partial \theta _{h}^{(L-1)} } } \\
{{\mathop{}\nolimits^{}} {\mathop{}\nolimits^{}} {\mathop{}\limits^{}} =-\eta \frac{\partial E}{\partial \beta _{qj} } \frac{\partial \beta _{qj} }{\partial z_{qj} } \frac{\partial z_{qj} }{\partial \alpha _{hj} } \frac{\partial \alpha _{hj} }{\partial y_{hj}^{(L-1)} } } \\ {{\mathop{}\nolimits_{}} {\mathop{}\nolimits_{}} {\mathop{}\limits^{}} =-\eta \delta _{h}^{(L-1)} }
\end{equation}

Note that, \begin{equation}
\delta _{h}^{(L-1)} =-\frac{\partial E}{\partial \beta _{qj} } \frac{\partial \beta _{qj} }{\partial z_{qj} } \frac{\partial z_{qj} }{\partial \alpha _{hj}^{(L-1)} } \frac{\partial \alpha _{hj}^{(L-1)} }{\partial y_{hj}^{(L-1)} } =\sigma ^{'} (y_{hj}^{(L-1)} )\cdot \sum _{q=1}^{Q}w_{qh}^{(L-1)} \cdot  \delta _{q}^{(L)}
\end{equation}

Here, H is the neurons' number in the L-1, P is the neurons' number in the L-2 layer, $\Delta w_{hp}^{(L-2)} $ is the weight change value from the \emph{p}th (with p=1,2, . . . , P) neuron of the L-2 layer to the \emph{h}th (with h=1,2,..,H) neuron of the L-1 layer, $\Delta \theta _{h}^{(L-1)} $ is the bias change value of the \emph{h}th (with h=1,2,..,H) neuron in the L-1 layer . $\sigma (\cdot )$ is the activation function. $\alpha_{pj}^{(L-1)} = \sigma (y_{hj}^{(L-1)} )= \sigma ($$\sum _{p=1}^{P}w_{hp}^{(L-2)} \alpha _{pj}^{(L-2)} +\theta _{h}^{(L-1)}  )$, $\alpha ^{(L-1)} $ is the activation value of neurons in the L-1 layer, $\alpha ^{(L-2)} $ is the activation value of neurons in the L-1 layer.

Thus, the weights and biases between the L-1 layer and the L-2 layer are updated as,
\begin{equation}
w_{hp}^{(L-2)} (g+1)=w_{hp}^{(L-2)} (g)+\Delta w_{hp}^{(L-2)}
\end{equation}
\begin{equation}
\theta _{h}^{(L-1)} (g+1)=\theta _{h}^{(L-1)} (g)+\Delta \theta _{h}^{(L-1)}
\end{equation}
Through the above derivation, we can generalize the general update formulas for the L-l layer to the L-(l-1)（with l=1,2,..,L-2） layer weights and biases, can be written as below:
\begin{equation}
\delta _{}^{(L-l)} =\sigma ^{'} (y_{}^{(L-l)} )\cdot \sum _{}^{}w_{}^{(L-l)} \cdot  \delta _{}^{(L-l+1)}
\end{equation}
then,
\begin{equation}
\Delta w^{(L-l)} =\eta \tau (x)[\delta ^{(L-l)} \alpha ^{(L-l+1)} -\lambda w^{(L-l+1)} ]
\end{equation}
\begin{equation}
\Delta \theta ^{(L-l+1)} =-\eta \delta ^{(L-l+1)}
\end{equation}
Thus, the weights and biases are updated as
\begin{equation}
w^{(L-l)} (g+1)=w^{(L-l)} (g)+\Delta w^{(L-l)}
\end{equation}
\begin{equation}
\theta ^{(L-l+1)} (g+1)=\theta ^{(L-1)} (g)+\Delta \theta ^{(L-l+1)}
\end{equation}

\section{Experiments}

\subsection{Overview of the Datasets}
Ten classification benchmark datasets is employed to evaluate models. Table 1 describes the characteristics of the datasets, containing datasets name (Data set), abbreviation of the datasets (Abbr.), size of the datasets (Size), dimensions of the datasets (Dim.), and number of classes (Class).
\begin{table}
\centering
\caption{Datasets Description}
\begin{tabular}{p{6cm} p{1.7cm} p{1.3cm} p{1.2cm}  p{1.2cm} }
\hline
Data set  &  Abbr. & Size &  Dim. & Class\\
\hline
Pima Indians Diabetes		&		Diabetes 	&		392 	&		8 &		   2\\
Congressional Voting Records	&		Vote		&		232 	&		16	&	   2\\
Risk Factors Cervical Cancer	&		RFCC	&		668		&	33 		 &  2 \\
SPECT-heart 		&				SPECT		&	267 &			22		  & 2 \\
Climate Model Simulation Crashes &		CMSC	&		540 &			18 	&	   2  \\
Website Phishing 		&			Web		&		1353	 &		9	&	   3 \\
Hayes-Roth			&			HR		&		160	&		 3 	&	   3\\
Balance Scale		&				Balance	&		625 &			4 &	   3 \\
Wine			&				Wine	&		178 	&		13       &   3\\
User Knowledge Modeling	&		UKM		&	403 	&		5 	&	   4 \\
\hline
\end{tabular}
\end{table}

\subsection{Comparison Methods}
We choose different types of classifier in the experiment to compare the efficiency with the GDFC method. Table 2 is used to introduce these methods. For a fair comparison with the proposed method, the potentiality of these methods is explored, and their parameters are tuned by trial and error.

\begin{table}
\centering
\caption{Comparative Methods}
\begin{tabular}{p{5cm} p{6cm} }
\hline
Type  &  Method\\
\hline
{Neural Network-based Methods} & Feed-forward Neural Network (FNN)\\& Nearest Neighbor Partitioning (NNP)\\& Floating Centroid Method (FCM)\\
\hline
{Other Classification Methods}  & Naïve Bayes (NB)\\ &Support Vector Machine (SVM)\\ & K-nearest Neighbor (KNN)  \\
\hline
\end{tabular}
\end{table}

\begin{itemize}
  \item KNN is a classical classification method. For the KNN, the number of nearest neighbors is selected in the range \{1, 30\}.

  \item  For the SVM, cost parameter is selected from the range $\{2^{-2},2^{-5}\}$.

  \item For the GDFC and FCM, the number of hidden layer neurons is set from range \{1, 40\}. the number of dimensions of the partition space is selected from \{N, 10N\}, and for the number of centroids is from the range \{N, 5N\}, where N is equal to the number of classes.

  \item For NNP, hidden layer neurons number is selected from the range \{1, 40\}. The dimension of partition space is chosen from the range \{N, 10N\}, and the number of centroids from the range {N, 5N}. The value of parameter p fixed at 3.

  \item For FNN, hidden layer neurons number is selected from the range \{1, 40\}.
\end{itemize}

\subsection{Evaluation Metrics}
As the evaluation metrics, Generalization Accuracy (GA) and Average F-measure (Avg.FM) is adopted to evaluate the efficiency of the GDFC method.
\begin{equation}
 GA = \frac{TC}{TN} \times 100\%	
\end{equation}
\begin{equation}
Avg.FM = \frac{2 \times (\frac{precision \times recall}{precision + recall})}{Num} \times 100\%
\end{equation}	
 The performance of all methods is evaluated by using ten-fold cross-validation. At first, the whole dataset is randomly split into ten subsets and then one subset is chosen for testing, and other subsets are employed for training. This whole process repeated for ten times, and the final result is considered by the mean of ten results.

\begin{table}
\centering
\caption{Accuracy comparison with neural network-based methods. The unit of results is percentage.}
\begin{tabular}{p{3cm} p{2cm} p{2cm} p{2cm} p{1.2cm} }
\hline
 Method & FNN &	NNP &	FCM &	GDFC\\
\hline
Diabetics	 &		75.5	 &		75	 &		77.75	 &	\textbf{79.5}\\
Vote	 &		78.75	 &		92.9	 &		92.9	 &		\textbf{94.17}\\
RFCC	 &		88.82	 &		87.06		 &	90	 &		\textbf{92.35}\\
SPECT	 &	80.71	 &		78.21	 &		80.71	 &		\textbf{82.07}\\
CMSC	 &		92.55	 &		92.73	 &		94	 &		\textbf{96.09}\\
Web	 &		86.64	 &		84.5	 &		85.11		 &	\textbf{88.41}\\
HR	 &		71.11	 &		79.44	 &		77.22	 &		\textbf{81.11}\\
Balance	  &		95.08	 &		94.6	 &		95.87	 &		\textbf{96.35}\\
Wine	&		94.44	 &	\textbf{98.89}	 &	\textbf{98.89}		 &	\textbf{98.89}\\
UKM		&		89.05	 &		95.24	 &		95.95	 &		\textbf{96.18}\\
\hline
\textit{\textbf{MEAN}}		&	85.27	 &		87.86	 &		88.84	 &		\textbf{90.51}\\
\hline
\end{tabular}
\end{table}

\subsection{Results Analysis}
We demonstrate the experimental results and the findings in this subsection. In our experiments, we compared the proposed GDFC method with six different classifiers, including SVM, NB, KNN, FNN, NNP, and FCM. Table 3 and 4 exhibit the testing accuracy and Avg.FM of neural network-based methods on each dataset.

From Table 3, our proposed GDFC achieved better generalization accuracy on nine datasets out of ten. Only in Wine dataset, the generalaization performance of NNP, and FCM is alike to the proposed method. That phenomenon clarifies that the GDFC is promising compared with comparative methods in terms of generalization performance. Table 4 shows the comparison between the proposed method with neural network-based methods based on Average F-measure. As compared with other methods, GDFC has better performance on eight of the ten datasets. Only in Balance and Wine dataset FCM and NNP has slightly better performance respectively. That reveals GDFC has better generalization ability than other omparative methods. For classification task, Avg.FM is a significant appraisal for the classifiers because Avg.FM summaries both precision and recall to evaluate the performance. Which demonstrates that, the GDFC has a better balance between precision and recall compared to other competing methods.

Furthermore, based on mean accuracy which is the average results of each method on ten datasets. Gradient-based floating centroid method improved about \textbf{3.50\%} (versus FNN), \textbf{1.77\%} (versus NNP) and \textbf{1.15\%} (versus FCM) in terms of testing accuracy. Moreover, GDFC improved about \textbf{7.81\%} (versus FNN), \textbf{1.82\%} (versus NNP) and\textbf{ 3.33\%} (versus FCM) in terms of Avg.FM.
\begin{table}
\centering
\caption{Avg.FM comparison with neural neteork-based methods. The unit of results is percentage.}
\begin{tabular}{p{3cm} p{2cm} p{2cm} p{2cm} p{1.2cm} }
\hline
Method &	FNN &	NNP &	FCM &	GDFC\\
\hline
Diabetics & 	73.07	 & 73.25 & 	72.21 & \textbf{75.29}\\
Vote & 	82.69	 & 92.9	 &  92.9	 & \textbf{94.14}\\
RFCC & 	59.17 & 	77.23 & 	72.65 & 	\textbf{83.76}\\
SPECT & 	63.16 & 	68.12 & 	67.46 & 	\textbf{73.07}\\
CMSC & 	62.38 & 	77.95 & 	75.84 & 	\textbf{83.79}\\
Web & 	74.83 & 	81.4 & 	66.55	 & \textbf{85.14}\\
Hr & 	69.6 & 	80.46 & 	77.24 & 	\textbf{81.48}\\
Balance	  & 	84.87 & 	90.23 & 	\textbf{92.81} & 	89.96\\
Wine & 	93.6 & 	\textbf{98.97} & 	98.88 & 	98.93\\
UKM	 & 82.13 & 	94.88 & 	96.11 & 	\textbf{97.08}\\
\hline
\textit{\textbf{MEAN}}		&	74.55	 &		83.54	 &	81.27	 &		\textbf{86.27}\\
\hline
\end{tabular}
\end{table}
Fig. 2 and Fig.3 displays generalization accuracy and Avg.FM of other existing classification methods, including SVM, NB and KNN. GDFC method achieves better performance on ten datasets in terms of classification accuracy. For Avg.FM, GDFC achieves higher performance of eight datasets out of ten datasets. SVM achieves better performance on RFCC dataset, and in Wine dataset SVM has similar performance as GDFC. Furthermore, it is noticeable that the average accuracy of GDFC is higher than the other methods. Gradient-based floating centroid method improved about \textbf{2.83\%} (versus SVM), \textbf{4.24\%} (versus NB), \textbf{3.71\%} (versus KNN) and \textbf{ 4.02\%} (versus SVM), \textbf{6.58\%} (versus NB), \textbf{7.54\%} (versus KNN) in terms of Avg.FM. That concludes our proposed GDFC method is superior to the other competing methods.

\begin{figure}
\includegraphics[width=\textwidth]{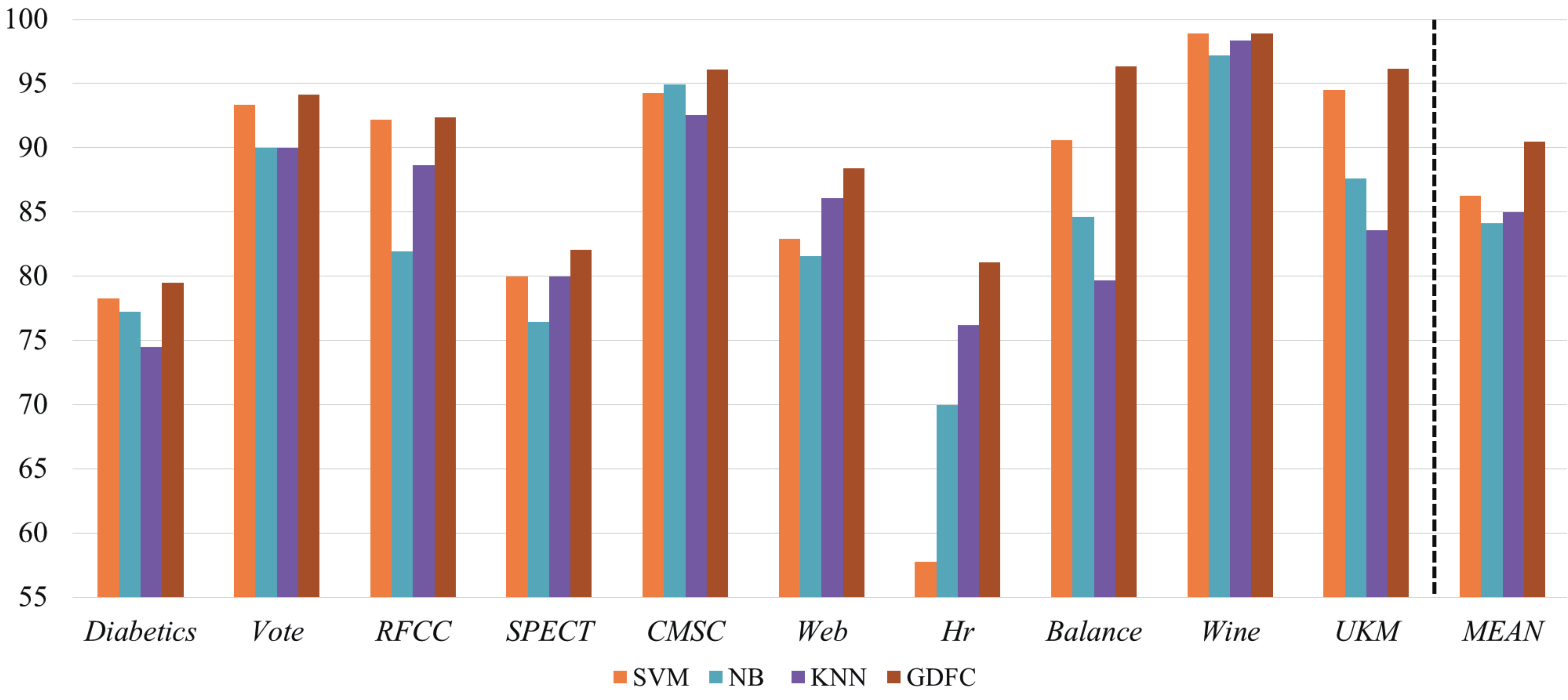}
\caption{Comparison between GDFC and the other classiﬁcation methods in terms of generalization accuracy.
}
\end{figure}

\begin{figure}
\includegraphics[width=\textwidth]{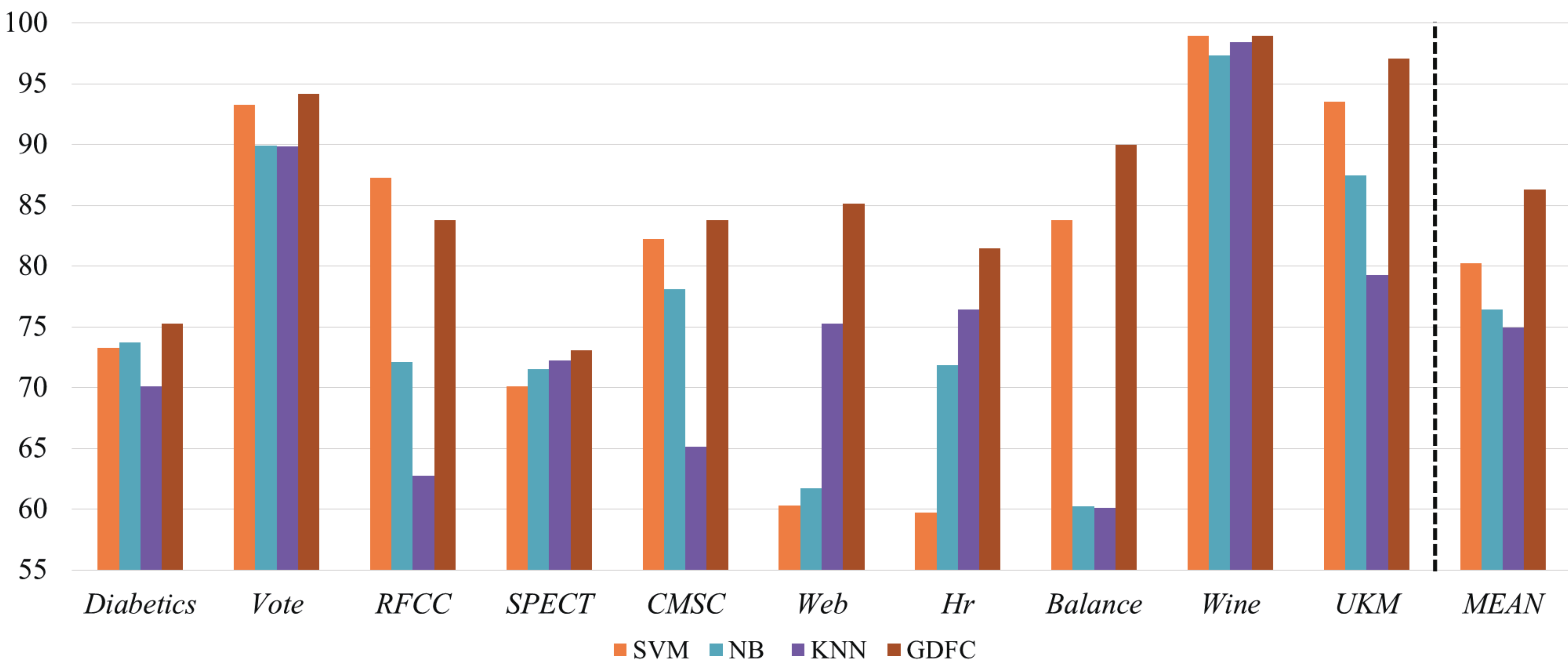}
\caption{Comparison between GDFC and the other classiﬁcation methods in terms of average F-measure.
}
\end{figure}


\section{Conclusions}
In this study, a novel gradient-based floating centroid (GDFC) method is introduced to solve the fixed-centroid problem for gradient-based neural network classifier. Moreover, the centroid loss function is introduced for maximizing the compactness of within-class and the separability of between-class during the optimization process. Experimental results indicated that proposed gradient-based floating centroid (GDFC) method outperformed the competing methods on the majority of the datasets in terms of classification accuracy and Avg.FM. In the future, different neural network structures will be investigated and employed in GDFC, considering that the mapping ability of the neural network is one of the vital factors for the classification performance of GDFC.

\section*{}
\textbf{Acknowledgements.} This work was supported by National Natural Science Foundation of China under Grant No. 61573166, No. 61572230, No. 61872419, No. 61873324, No. 81671785, No. 61672262. Project of Shandong Province Higher Educational Science and Technology Program under Grant No. J16LN07. Shandong Provincial Natural Science Foundation No. ZR2019MF040, No. ZR2018LF005. Shandong Provincial Key R\&D Program under Grant No. 2018GGX101048, No. 2016ZDJS01A12, No. 2016GGX101001, No. 2017CXZC1206. Taishan Scholar Project of Shandong Province, China.

\bibliographystyle{splncs04}
\bibliography{GDFC-backup}
\end{document}